\documentclass[journal]{IEEEtran}
\usepackage{amsmath,amsfonts}
\usepackage{algpseudocode}
\usepackage{algorithmicx,algorithm}
\usepackage{hyperref}
\usepackage{array}
\usepackage[caption=false,font=normalsize,labelfont=sf,textfont=sf]{subfig}
\usepackage{caption}
\usepackage{textcomp}
\usepackage{stfloats}
\usepackage{url}
\usepackage{verbatim}
\usepackage{graphicx}
\usepackage{cite}

\usepackage[square,sort&compress,numbers]{natbib}
\usepackage{amsmath}
\usepackage{amsfonts} 
\usepackage{amssymb} 
\usepackage{multirow}
\usepackage{threeparttable}
\usepackage{booktabs}
\usepackage{amsmath}
\usepackage{bbding}
\usepackage{xcolor}
\usepackage{soul}
\usepackage{pifont}

\graphicspath{{figs/}}

\definecolor{emb_a}{HTML}{cfe3f3}
\definecolor{emb_b}{HTML}{ffcec3}
\definecolor{blue}{HTML}{2a00fe}
\definecolor{red}{HTML}{ff0300}

\begin{document}

\title{CCExpert: Advancing MLLM Capability in Remote Sensing Change Captioning with Difference-Aware Integration and a Foundational Dataset
}

% \author{
% Mingze~Wang$^{1,\dag}$,~Keyan~Chen$^{1,\dag}$,~Lili~Su$^2$,~Cilin~Yan$^1$,~Pengcheng~Yuan$^3$,~Xiaolong~Jiang$^3$,\\~Baochang Zhang$1,^\star$
% ~\IEEEmembership{Member,~IEEE},~and~Jinhu Lv$^1$~\IEEEmembership{Fellow,~IEEE}

\author{
Zhiming~Wang$^{1\dagger}$,~Mingze~Wang$^{1\dagger}$,~Sheng~Xu$^{1}$,~Yanjing~Li$^{1}$,~and~Baochang~Zhang$^{1\star}$
\vspace{6pt}
\\
% Beihang University$^1$,~ByteDance$^2$
Beihang University$^1$

\thanks{$\dagger$: Equal contribution.}
\thanks{$\star$: Corresponding Author.}
\thanks{This work is currently in progress(WIP), with ongoing development and refinement.}
}
% $\dag$: Equal contribution, 

\markboth{Journal of \LaTeX\ Class Files,~Vol.~14, No.~8, August~2015}%
{Shell \MakeLowercase{\textit{et al.}}: Bare Demo of IEEEtran.cls for IEEE Journals}

\maketitle

\begin{abstract}

Remote Sensing Image Change Captioning (RSICC) aims to generate natural language descriptions of surface changes between multi-temporal remote sensing images, detailing the categories, locations, and dynamics of changed objects (e.g., additions or disappearances). Many current methods attempt to leverage the long-sequence understanding and reasoning capabilities of multimodal large language models (MLLMs) for this task. However, without comprehensive data support, these approaches often alter the essential feature transmission pathways of MLLMs, disrupting the intrinsic knowledge within the models and limiting their potential in RSICC. In this paper, we propose a novel model, CCExpert, based on a new, advanced multimodal large model framework. Firstly, we design a difference-aware integration module to capture multi-scale differences between bi-temporal images and incorporate them into the original image context, thereby enhancing the signal-to-noise ratio of differential features. Secondly, we constructed a high-quality, diversified dataset called CC-Foundation, containing 200,000 image pairs and 1.2 million captions, to provide substantial data support for continue pretraining in this domain. Lastly, we employed a three-stage progressive training process to ensure the deep integration of the difference-aware integration module with the pretrained MLLM. CCExpert achieved a notable performance of $S^*_m=81.80$ on the LEVIR-CC benchmark, significantly surpassing previous state-of-the-art methods. The code and part of the dataset will soon be open-sourced at \url{https://github.com/Meize0729/CCExpert}.
\end{abstract}


\IEEEpeerreviewmaketitle
\section{Introduction}

\begin{figure}[t]
\begin{center}
\includegraphics[width=0.75\linewidth]{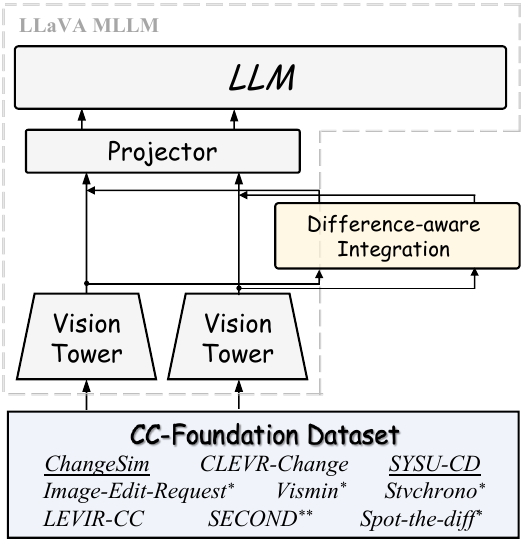}
\end{center}
\caption{
Basic components of the proposed MLLM-based CCExpert and the CC-Foundation Dataset. * indicates that the dataset has undergone rewriting and refinement, underline denotes datasets generated by GPT-4o, and ** indicates datasets that have been expert-annotated and refined.
}
\label{fig:CCExpert_first}
\end{figure}

\IEEEPARstart{R}emote sensing image change analysis is a pivotal technique for monitoring dynamic Earth processes, relying on comparing images of the same region taken at different times. It is widely applied in environmental monitoring and disaster management. With advancements in remote sensing and Earth observation technologies, Remote Sensing Image Change Captioning (RSICC) has emerged as an innovative method combining natural language processing and computer vision to generate natural language descriptions that explain semantic changes between bitemporal remote sensing images. RSICC not only provides precise and insightful descriptions for understanding dynamic land cover changes but also serves as a powerful tool in fields like urban planning and environmental monitoring~\cite{Semantic-CC,Diffusion-RSCC,RSCaMa}.

Early RSICC methods \cite{RSICCformer, PSNet, PromptCC, Sen, SFT} primarily used CNN or Transformer-based encoder-decoder frameworks. These models incorporated techniques such as hierarchical self-attention modules, complex bi-temporal feature interaction modules, and multi-task joint learning \cite{Pix4Cap, Semantic-CC} to address challenges in RSICC, including difficulty in bi-temporal image feature alignment and loss of detail.

However, these smaller, non-pretrained models often struggle with extensive and high-noise contexts, failing to capture precise semantic changes. With advancements in foundation models \cite{sam, rsprompter, rsbuilding}, large language models (LLMs) \cite{yang2024qwen2, achiam2023gpt}, and multimodal learning \cite{clip}, large multimodal models (MLLMs) \cite{Qwen-VL, minigpt4, liu2023llava, li2024llava, liu2024llavanext}, pre-trained on extensive, aligned datasets, have shown remarkable success in cross-modal tasks such as image captioning and visual question answering. In change captioning, the use of MLLMs with powerful generalization capabilities has become increasingly prevalent, achieving significant results through targeted architectural adjustments and supervised fine-tuning (SFT) on domain-specific datasets.

In the field of natural image Change Caption, researchers have enhanced the model's ability to accurately capture semantic changes through various approaches, such as module design~\cite{VIR-VLFM}, architectural adjustments, data construction~\cite{OneDiff}, and retrieval augmentation~\cite{ZWW2024}. These efforts have also improved the comprehensiveness and precision of descriptions, providing new directions and methodologies for the advancement of RSICC.

Similarly, in the RSICC field, researchers have moved away from complex end-to-end encoder-decoder architectures, instead adopting the MLLM paradigm to derive numerous innovative approaches that have demonstrated excellent performance. Specifically, Tsujimoto et al. proposed a prompt-based multimodal model that significantly improved the generation of change descriptions for bi-temporal remote sensing images through stepwise reasoning. Zhu et al. developed Semantic-CC~\cite{Semantic-CC}, which integrates the SAM visual encoder~\cite{sam} and MiniGPT4~\cite{minigpt4}, and employs pixel-level semantic guidance for multitask learning to enhance the accuracy and robustness of change descriptions. Noman et al. introduced CDChat~\cite{CDChat}, which fine-tunes the MLLM using the SYSU-CD and LEVIR-CD datasets. Deng et al. developed ChangeChat~\cite{ChangeChat}, which enhances interaction and contextual understanding in bi-temporal visual language models through multimodal instruction tuning and the ChangeChat-87k dataset. Yang et al. proposed the KCFI~\cite{KCFI} framework, which leverages visual instructions and pixel-level change detection tasks to introduce key change features, effectively filtering irrelevant features.

However, most existing MLLM-based RSICC methods, lacking comprehensive data support, modify the core structure of multimodal large models or disrupt essential feature transmission pathways, thereby undermining the intrinsic knowledge within these models. For example, Semantic-CC replaces the traditional CLIP-pretrained visual tower with a SAM~\cite{sam} backbone and freezes the Q-Former component~\cite{li2023blip}, limiting the expression of visual features. Similarly, KCFI inputs specially processed change features instead of raw visual features and lacks further pretraining, which restricts the model’s understanding of these features. Although these methods achieve certain results, they fall short of fully leveraging the potential of MLLMs in RSICC.

Therefore, this paper aims to enhance MLLM performance in the RSICC domain by preserving the inherent knowledge system within MLLMs as much as possible, while using data-driven continue pretraining and specially designed modules.

In view of the significant progress in multimodal large models in recent years, particularly in tasks involving image-text interaction, multi-image understanding, and reasoning, this paper initially explores various base models, including MiniGPT-4 \cite{minigpt4}, LLaVA-1.5~\cite{liu2023llava}, and LLaVA-OneVision~\cite{li2024llava}. Among these, LLaVA-OneVision achieved a strong performance on the LEVIR-CC dataset with a supervised fine-tuning (SFT) approach, yielding an $S^*_m=80.19$ without altering the model architecture. Consequently, subsequent experiments in this study will be based on LLaVA-OneVision, utilizing the most straightforward paradigm: independently encoding dual-temporal images, followed by large language model (LLM) analysis of image features to generate descriptive outputs. This approach allows for maximal retention of the pretrained knowledge embedded in the base model.

Inspired by works such as VIR-VLFM, Semantic-CC~\cite{VIR-VLFM, Semantic-CC}, we designed a Difference-aware Integration Module, comprising a Difference Enhancement submodule (Diff Expert) and a Multi-Scale Adaptive Weighting submodule. This module is intended to explicitly capture multi-scale differences within images and integrate them into the original image features, thereby enhancing the representation of differential features.

Additionally, to enhance CCExpert's generalization and foundational capabilities in complex scene understanding, we constructed a large-scale dataset, the CC-Foundation Dataset, comprising 200,000 image pairs and 1.2 million captions for continued pretraining. The CC-Foundation Dataset consists of three parts: (1) a collection of multiple open-source change captioning datasets within the field, refined and optimized using large language models; (2) an expansion of data based on Change Detection datasets, where ground-truth change masks serve as prompts to generate change descriptions through GPT-4o; and (3) an additional set of semantically rich datasets annotated by domain experts, further enhancing the dataset’s diversity and challenge level. Finally, we implemented a three-stage training process to ensure the deep integration of the differential injection module with the existing multimodal large model, achieving significant improvements across multiple evaluation metrics.

In summary, we propose CCExpert, an expert model specifically designed for remote sensing image change captioning, as illustrated in Fig.~\ref{fig:CCExpert_first}. This model leverages an advanced multimodal large model as its foundation, integrates a carefully designed Difference-aware Integration Module, and undergoes staged training based on the large-scale CC-Foundation Dataset. 

\vspace{4pt}
The primary contributions of this paper can be summarized as follows:

i) We utilize the latest, state-of-the-art base model and design a Difference-aware Integration Module to capture and enhance fine-grained differential features between image pairs. This module injects extracted change information directly into the image features without altering the core model architecture, facilitating easier interpretation of differences by the LLM.

ii) We constructed the CC-Foundation Dataset, comprising 200,000 pairs of remote sensing images and 1.2 million captions. This dataset integrates and optimizes multiple open-source datasets, further expanded through multimodal large models and domain expert annotations to enhance both scale and diversity. The dataset provides rich support for remote sensing change captioning, significantly boosting the model’s foundational capabilities.

iii) We designed a three-stage training process, progressively fine-tuning CCExpert to optimize its performance. This approach ensures a deep integration of the Difference-aware Integration Module with the MLLM. Experimental results demonstrate that CCExpert achieves substantial performance improvements on the LEVIR-CC benchmark.
\vspace{4pt}

The remainder of this paper is organized as follows: Sec. II presents a comprehensive review of the relevant literature. In Sec. III, we delve into the specifics of the structure and training strategy of our proposed CCExpert. Sec. IV introduces the datasets and the evaluation protocols and provides an in-depth analysis of both quantitative and qualitative results. This section further includes ablation studies, supplemented by a thorough discussion and an acknowledgment of limitations. Lastly, Sec. V concludes the paper by encapsulating the key findings and insights.

\section{Related Works}

\subsection{Multimodal Large Language Model}

Inspired by the powerful reasoning capabilities of large language models (LLMs), researchers are actively exploring ways to integrate these abilities into the visual domain, thereby advancing the development of multimodal LLMs (MLLM). Proprietary multimodal models such as GPT-4V~\cite{openai2023gpt4}, GPT-4o~\cite{openai2024gpt4o}, Gemini~\cite{team2023gemini}, and Claude-3.5~\cite{anthropic2024claude35} have demonstrated outstanding performance across a wide range of tasks. Simultaneously, the open-source community has produced a steady stream of innovative developments. For example, Flamingo~\cite{flamingo} employs cross-attention mechanisms to facilitate visual context learning, while BLIP-2~\cite{li2023blip} and mPLUG-OWL~\cite{ye2023mplug} use visual encoders to input image features alongside text embeddings into LLMs. Otter~\cite{li2023otter} enhances few-shot learning capabilities through context instruction tuning on the MIMIC-IT dataset, while LLaVA~\cite{liu2023llava} and MiniGPT-4~\cite{minigpt4} first align image-text features before applying instruction fine-tuning. Additionally, models like VisionLLM~\cite{wang2023visionllm}, Kosmos-2~\cite{peng2023kosmos}, and DetGPT~\cite{detgpt} demonstrate significant flexibility in multi-task interactions, including object detection and image segmentation.

With growing interest and active research in this field, several recent open-source models have started to rival—and even surpass—some proprietary models. For instance, LLaVA-NeXT~\cite{liu2024llavanext} enhances the LLaVA series through a dynamic resolution input strategy and new data utilization. DeepSeek-VL~\cite{lu2024deepseek} combines a hybrid visual encoder with innovative training strategies to deliver competitive results, while the InternVL series~\cite{Chen2023InternVS,Chen2024HowFA} consistently excels in benchmark tests thanks to high-quality, multi-source image-text data. The latest LLaVA-OneVision~\cite{li2024llava} has become the first open-source multimodal model to push performance limits across three major computer vision tasks: single-image, multi-image, and video processing. Furthermore, Qwen2-VL~\cite{wang2024qwen2} leverages a robust base model, dynamic resolution mechanisms, and multimodal rotational position embeddings to process images of varying resolutions dynamically. This approach effectively integrates spatial information from text, images, and video, achieving exceptional performance metrics.

\subsection{Remote Sensing Image Change Caption}

Image change captioning lies at the intersection of computer vision and natural language processing (NLP) and has attracted significant attention in recent years. This section briefly reviews progress in this area within computer vision and remote sensing.

Jhamtani et al.\cite{HT2018} were the first to introduce a dataset and model for image change captioning, employing latent variables to model visual salience. Park et al.\cite{PDR2019} developed a dual dynamic attention model (DUDA) to separate irrelevant interference from semantic changes. Kim et al.\cite{KKL2021} proposed a viewpoint-agnostic change captioning network (VACC) that captures genuine changes through a difference encoder and a cycle consistency module. Shi et al.\cite{SYG2020} designed a novel visual encoder to distinguish between viewpoint and semantic changes, optimizing the attention mechanism with reinforcement learning. Hosseinzadeh et al.\cite{HW2021} introduced auxiliary tasks to improve captioning accuracy, while Zheng et al.\cite{ZMS2022} designed a role-playing dialogue system to identify differences between images. FINER-MLLM~\cite{ZWW2024} employs LoRA fine-tuning, dual constraints, and retrieval enhancement to accurately describe subtle changes.

In the remote sensing field, Chouaf et al.\cite{CHS2021} were the first to apply captioning techniques to bi-temporal remote sensing images. Hoxha et al.\cite{HCM2022} and Liu et al.\cite{RSICCformer} improved captioning accuracy by introducing new datasets and adopting models based on RNN or SVM and Transformers, respectively. Liu et al.\cite{PromptCC} used a multi-prompt learning strategy, dividing the task into change detection and description. Chang et al.\cite{Chg2Cap} introduced the Chg2Cap network, which combines a Siamese CNN, attention encoder, and Transformer generator to effectively mitigate the effects of lighting and seasonal changes. Zhou et al.\cite{Sen} proposed the Single-Stream Extractor Network (SEN), integrating shallow feature embedding and cross-attention modules to enhance spatiotemporal feature representation. Liu et al.~\cite{RSCaMa} introduced the RSCaMa model, employing a state-space module to significantly improve the efficiency and accuracy of remote sensing change descriptions.

Early RSICC methods often employed CNN or Transformer-based encoder-decoder architectures, utilizing techniques such as hierarchical self-attention modules, dual-temporal feature interaction modules, and multi-task learning~\cite{Pix4Cap, Semantic-CC} to tackle challenges like feature comparison and detail retention. With advancements in foundation models~\cite{sam, rsprompter, rsbuilding}, large language models (LLMs)\cite{yang2024qwen2, achiam2023gpt}, and multimodal learning\cite{clip}, multimodal large models (MLLMs) have achieved impressive results in cross-modal tasks such as image captioning and visual question answering through extensive data alignment and pretraining. Change captioning has increasingly leveraged MLLMs, adapting their robust generalization capabilities and fine-tuning on specific data through supervised fine-tuning (SFT), with notable success.

In the natural image change captioning domain, Lu et al. introduced VIR-VLFM~\cite{VIR-VLFM} based on Instruct-BLIP, adding trainable adapters, viewpoint alignment flow, and semantic emphasis modules to address limitations from single-image understanding and viewpoint variations. Subsequently, Hu et al. proposed OneDiff~\cite{OneDiff}, integrating a twin image encoder and visual difference module with a two-stage training strategy and a new DiffCap dataset to achieve fine-grained difference detection and description. Recently, FINER-MLLM~\cite{ZWW2024} has used LoRA fine-tuning, dual constraints, and retrieval enhancement to precisely describe subtle changes.

In RSICC, Tsujimoto et al. introduced a prompt-based multimodal model to improve the generation of bi-temporal remote sensing image captions through stepwise reasoning. Zhu et al. developed Semantic-CC~\cite{Semantic-CC}, which combines the SAM visual encoder~\cite{sam} and the MiniGPT4 language model~\cite{minigpt4} and uses pixel-level semantic guidance for multi-task learning to enhance caption accuracy and robustness across temporal and spatial scenarios. Noman et al. introduced CDChat~\cite{CDChat}, fine-tuning MLLMs on annotated SYSU-CD~\cite{shi21deeply} and LEVIR-CD~\cite{chen2020spatial} datasets. Deng et al. developed the ChangeChat~\cite{ChangeChat} model, optimizing bi-temporal visual language models through multimodal instruction tuning and the ChangeChat-87k dataset to enhance interactivity and contextual understanding. Yang et al. proposed the KCFI~\cite{KCFI} framework, which incorporates visual instructions and pixel-level change detection tasks, guiding the model with key change features and effectively filtering irrelevant information.

\subsection{Summary}

Currently, MLLM-based RSICC is still in its early stages. Many existing methods fail to treat the MLLM as a well-pretrained whole, instead isolating each module as an independent component, overly focused on architectural innovations without fully leveraging the pretrained knowledge structure. Additionally, current approaches lack sufficient attention to data utilization, limiting their ability to address the potential knowledge bias arising from discrepancies between the pretraining and application domains of MLLMs.

Thus, our objective is to maximize retention of the MLLM's inherent knowledge structure while using data-driven continued pretraining to correct potential knowledge biases. Through specially designed modules, we explicitly enhance visual context features, thereby strengthening the performance of MLLMs in the RSICC domain.

\section{Methodology}

\begin{figure*}[!thb]
\begin{center}
\includegraphics[width=0.9\linewidth]{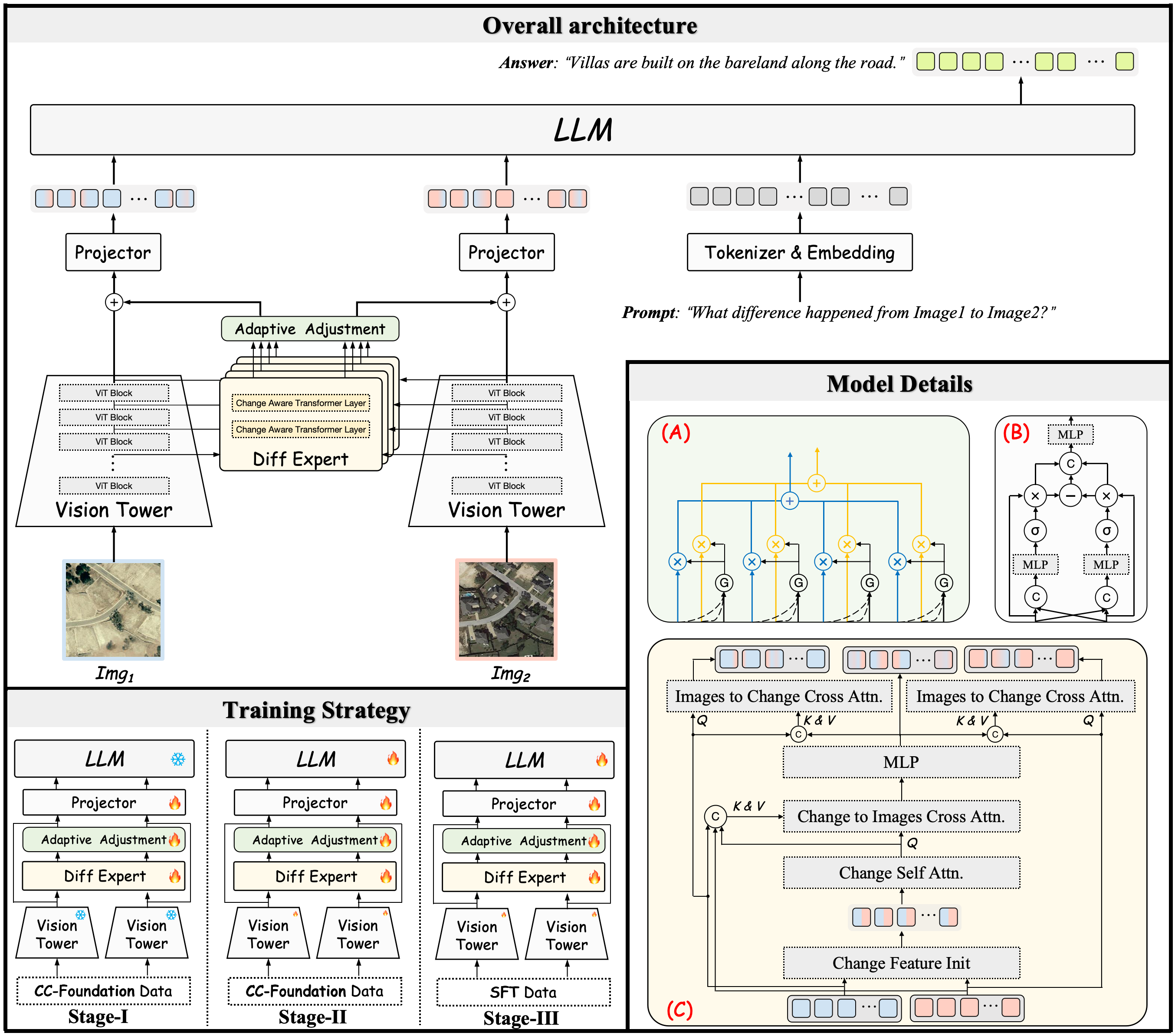}
\end{center}
\caption{
The overall architecture, component details, and training strategy of CCExpert. Model Details(A): Adaptive Adjustment module; Model Details(B): Change Feature Init module in Change Aware Transformer Layer; Model Details(C): Change Aware Transformer Layer.
}
\label{fig:CCExpert_overall_arch}
\end{figure*}

% >>>>>>>>>>>>>>>>>>>>>>>>>>>>>>>>> 总览 <<<<<<<<<<<<<<<<<<<<<<<<<<<<<<<<<<
This paper introduces CCExpert, an innovative approach for generating change captions for remote sensing images. CCExpert follows the basic architecture of multimodal large models, where temporal images are independently input and encoded. Next, a Difference-aware Integration Module injects change information to enhance contextual image features, and finally, the robust LLM interprets these differences and generates captions. The overall structure is illustrated in Fig.\ref{fig:CCExpert_overall_arch}, comprising four key components: an encoder for image feature extraction, a Difference-aware Integration Module to explicitly extract and integrate bi-temporal feature differences, a linear projection to map image features to the text space, and a large language model for change interpretation and caption generation. The image encoder, linear projection, and language model adhere to the basic configuration of LLaVA-OneVision\cite{li2024llava}. The process can be described as follows:

\begin{equation}
\begin{aligned}
F_1, F_2 &= \Phi_{\text{enc}} ( \mathcal{I}_1, \mathcal{I}_2) \\
{F_1}^\prime, {F_2}^\prime &= \Phi_{\text{enhancer}} (F_1, F_2) \\
\hat{F}_1, \hat{F}_2 &= \Phi_{\text{projector}} ({F_1}^\prime, {F_2}^\prime) \\
T &= \Phi_{\text{llm-decode}} ( \hat{F}_1, \hat{F}_2, E) \\
\end{aligned}
\end{equation}

Here, the images at different timestamps ($\mathcal{I}1 \in \mathbb{R}^{H \times W \times 3}$ and $\mathcal{I}2 \in \mathbb{R}^{H \times W \times 3}$) are first encoded by the encoder $\Phi{\text{enc}}$ to obtain bi-temporal features ($F_1 \in \mathbb{R}^{\frac{H}{14} \times \frac{W}{14} \times d}$ and $F_2 \in \mathbb{R}^{\frac{H}{14} \times \frac{W}{14} \times d}$). Next, these features are fed into the module $\Phi{\text{enhancer}}$, which performs multi-scale difference feature interaction and aggregation to generate change-sensitive image features ${F_1}^\prime \in \mathbb{R}^{\frac{H}{14} \times \frac{W}{14} \times d}$ and ${F_2}^\prime \in \mathbb{R}^{\frac{H}{14} \times \frac{W}{14} \times d}$. These enhanced features are then passed through the module $\Phi_{\text{projector}}$ to convert them from the image space to the text space, yielding LLM-compatible features $\hat{F}_1 \in \mathbb{R}^{\frac{H}{14} \times \frac{W}{14} \times c}$ and $\hat{F}2 \in \mathbb{R}^{\frac{H}{14} \times \frac{W}{14} \times c}$. Finally, prompts are designed, tokenized, and embedded to produce $E$, which is concatenated with the image features. Leveraging the understanding and text generation capabilities of the large language model $\Phi{\text{llm-decode}}$, we obtain the change description text ($T \in \mathcal{C}^{N}$, where $\mathcal{C}$ represents the vocabulary and $N$ is the description length).

% @@@@@@@@@@@@@@@@@@@@@@@@@@@@@@@@ 总览End @@@@@@@@@@@@@@@@@@@@@@@@@@@@@@@@

% >>>>>>>>>>>>>>>>>>>>>>>>>>>>>>>>> Vision Tower <<<<<<<<<<<<<<<<<<<<<<<<<<<<<<<<<<

\subsection{Vision Tower}
In this section, we describe the architecture of the image encoder in CCExpert. The image encoder, pretrained through image-text contrastive learning~\cite{clip} and multimodal large model pretraining, generates robust visual features. Initially, the bi-temporal image features are extracted independently, with no cross-interaction between the features from the two temporal images. The encoding process is summarized as follows:

\begin{equation}
\begin{aligned}
    \{ F_1^i \} = \Phi_{\text{enc}} (\mathcal{I}_1) \\
    \{ F_2^i \} = \Phi_{\text{enc}} (\mathcal{I}_2)
\end{aligned}
\end{equation}

CCExpert employs a Vision Transformer (ViT)\cite{vit} as the image encoder, a common choice among multimodal large models. ViT\cite{vit} segments the input image into non-overlapping patches of fixed size. Each patch is flattened and linearly projected into a fixed-length embedding. The relationships among these embeddings are modeled through stacked Transformer encoder layers, enabling the capture of global semantic information within the image. Specifically, CCExpert adopts the siglip-so400m architecture~\cite{zhai2023sigmoid}, which consists of 27 Transformer layers.

% @@@@@@@@@@@@@@@@@@@@@@@@@@@@@@@@ Vision Tower End @@@@@@@@@@@@@@@@@@@@@@@@@@@@@@@@

% >>>>>>>>>>>>>>>>>>>>>>>>>>>>>>>>> Diff <<<<<<<<<<<<<<<<<<<<<<<<<<<<<<<<<<

\subsection{Difference-aware Integration Module}

Multimodal large models (MLLMs) typically use visual features from the penultimate layer of the Vision Tower. However, significant differential features are often embedded in other feature levels. When the change regions in the images are not prominent at the feature level, relying solely on the contrastive understanding of the LLM may limit the accuracy of generated descriptions. To address this, we design a Difference-aware Integration Module that extracts comprehensive multi-scale change information from bi-temporal images, enhancing the prominence of differential features. This module comprises two sub-modules: "Diff Expert" and "Adaptive Adjustment."

First, the Diff Expert module extracts differential features from multi-scale bi-temporal features and injects them to reinforce the original image features. Next, the Adaptive Adjustment module dynamically weights the features at different scales, producing a multi-scale representation. Finally, residual connections combine the multi-scale features with the original features, ensuring that critical information is retained and forming the final image representation.

This design enables the model to identify subtle changes, utilizing multi-scale semantic information and details for precise and complete description generation.

\subsubsection{Diff Expert}

This sub-module independently processes multi-scale features from the bi-temporal images extracted by the image encoder, denoted as ${ F_1^i } \in \mathbb{R}^{\frac{H}{14} \times \frac{W}{14} \times d}$ and ${ F_2^i } \in \mathbb{R}^{\frac{H}{14} \times \frac{W}{14} \times d}$, with $i \in \{-11, -8, -5, -2\}$. Differential features across scales are captured independently without cross-scale interference. This sub-module consists of two Change Aware Transformer Layers, with each layer’s structure illustrated in Fig.~\ref{fig:CCExpert_overall_arch}, Model-Details(c). The following section provides a detailed introduction to the Change Aware Transformer Layer.

A core component, "Change Feature Init" ($\Phi_{\text{change-init}}$), is inspired by Semantic-CC~\cite{zhu2024semantic}. It initializes differential feature maps for the two temporal images as follows: 
\begin{equation}
\begin{aligned}
    {F_1^i}_{\text{init}} &= F_1^i \otimes \sigma( \Phi_{\text{proj-init}} (\Phi_{\text{cat}} (F_1^i, F_2^i) ) )  \\
    {F_2^i}_{\text{init}} &= F_2^i \otimes \sigma( \Phi_{\text{proj-init}} (\Phi_{\text{cat}} (F_2^i, F_1^i) ) )  \\
    \Delta F^i &= \Phi_{\text{Cat}} ( {F_1^i}_{\text{init}}, {F_2^i}_{\text{init}}, {F_1^i}_{\text{init}} - {F_2^i}_{\text{init}} ) \\
    \Delta F^i &= \Phi_{\text{proj}} ( \Delta F^i ) \\
\end{aligned}
\end{equation}
where $i$ denotes the feature map index, $\Phi_{\text{cat}}$ represents concatenation along the channel dimension, and $\Phi_{\text{proj-init}}$ maps the 2d-dimensional feature to d-dimension.

The resulting multi-scale initial differential features $\Delta F^i \in \mathbb{R}^{\frac{H}{14} \times \frac{W}{14} \times d}$ are then processed through self-attention and cross-attention layers in the Change Aware Transformer Layer to construct and inject differential features into the bi-temporal image representations. The Change Aware Transformer Layer computation is recursively defined as: 
\begin{equation}
\begin{aligned}
\Delta F^i &= \Phi_{\text{change-init}}(F_1^i, F_2^i) \\
\Delta F^i &= \Phi_{\text{S-attn}}(\Delta F^i) \\
\Delta F^i &= \Phi_{\text{C-attn}}(\Delta F^i, \Phi_{\text{Cat}}(F_1^i, F_2^i, \Delta F^i)) \\
\Delta F^i &= \Phi_{\text{mlp-proj}}(\Delta F^i) \\
\tilde{F}_1^i &= \Phi_{\text{C-attn}}(F_1^i, \Phi_{\text{Cat}}(F_1^i, \Delta F^i)) \\
\tilde{F}_2^i &= \Phi_{\text{C-attn}}(F_2^i, \Phi_{\text{Cat}}(F_2^i, \Delta F^i)) \\
\end{aligned}
\end{equation}
At the $i$-th layer, the bi-temporal image features $F_1^i \in \mathbb{R}^{\frac{H}{14} \times \frac{W}{14} \times d}$ and $F_2^i \in \mathbb{R}^{\frac{H}{14} \times \frac{W}{14} \times d}$ first pass through the $\Phi_{\text{change-init}}$ (Change Feature Init) module to generate the initial difference feature $\Delta F^i \in \mathbb{R}^{\frac{H}{14} \times \frac{W}{14} \times d}$. Next, $\Delta F^i$ undergoes initial self-interaction through the multi-head self-attention module $\Phi_{\text{S-attn}}$ (Change Self Attention). Subsequently, with $\Delta F^i$ as the query, it engages in further interaction with the bi-temporal features $F_1^i$ and $F_2^i$ through the cross-attention module $\Phi_{\text{C-attn}}$ (Change to Images Cross Attention) to enhance differential feature extraction. Finally, the feature differences are processed through the linear projection layer $\Phi_{\text{mlp-proj}}$ (MLP) to yield the final feature difference $\Delta F^i$. Additionally, an "Images to Change Cross Attention" module is introduced, where $F^i$ serves as the query and interacts with the differential information $\Delta F^i$ through $\Phi_{\text{C-attn}}$, injecting change information into the image features to strengthen the representation of the bi-temporal features.

Ultimately, through the Diff Expert module composed of two Change Aware Transformer Layers, the differential features are fully expressed and guide the enhancement of the original image features. This results in multi-scale bi-temporal image features enriched with change information, ${\tilde{F}_1^i}$ and ${\tilde{F}_2^i}$, as well as the differential feature representations ${\Delta F^i}$.

\subsubsection{Adaptive Adjustment}

After capturing multi-scale bi-temporal image features with change information and their differential representations, directly inputting all features to the LLM for decoding would lead to excessive computational complexity. Additionally, the long sequence length poses a significant challenge to the LLM’s capacity. To address this, we designed a Multi-Scale Adaptive Weighting and Adjustment module that consolidates multi-scale information. This module dynamically computes weights for each scale, normalizes them, and then performs weighted summation. The final output is a bi-temporal image feature representation aggregated across multiple scales. This representation is directly added to the original image features to balance change information with the original semantic context. The process is recursively defined as follows:
\begin{equation}
\begin{aligned}
G^i &= \Phi_{\text{Mean}}(\Phi_{\text{Cat}}(\tilde{F}_1^i, \tilde{F}_1^i, \Delta F^i)) \\
Score^i &= \sigma \circ \Phi_{\text{Proj-1}} \circ \Phi_{\text{GELU}} \circ \Phi_{\text{Proj-2}}(G^i) \\
{F_1^i}^\prime &= Score^i * \tilde{F}_1^i \\
{F_2^i}^\prime &= Score^i * \tilde{F}_2^i \\
{F_1}^\prime &= \sum_{i} {F_1^i}^\prime + F_1^{k} \\
{F_2}^\prime &= \sum_{i} {F_2^i}^\prime + F_2^{k}\\
\end{aligned}
\end{equation}
Here, $i$ denotes the feature map index, $\Phi_{\text{cat}}$ represents concatenation along the channel dimension, $\Phi_{\text{Mean}}$ indicates average pooling along the token dimension, $\Phi_{\text{Proj-1}}$ and $\Phi_{\text{Proj-2}}$ are linear projections, and $\Phi_{\text{GELU}}$ represents the GELU activation function.

% @@@@@@@@@@@@@@@@@@@@@@@@@@@@@@@@ DIFF End @@@@@@@@@@@@@@@@@@@@@@@@@@@@@@@@

% >>>>>>>>>>>>>>>>>>>>>>>>>>>>>>>>> Projector <<<<<<<<<<<<<<<<<<<<<<<<<<<<<<<<<<

\subsection{Projector}

CCExpert uses a Projector to map features from image space to text space. For this, we adopt the basic Projector configuration from the Llava series of multimodal models~\cite{li2024llava}. The process can be expressed as follows:

\begin{equation}
\begin{aligned}
\hat{F}_1 &= \Phi_{\text{Proj-2}} \circ \Phi_{\text{GELU}} \circ \Phi_{\text{Proj-1}}({F_1}^\prime) \\
\hat{F}_2 &= \Phi_{\text{Proj-2}} \circ \Phi_{\text{GELU}} \circ \Phi_{\text{Proj-1}}({F_2}^\prime) \\
\end{aligned}
\end{equation}

In this configuration, $\Phi_{\text{Proj-1}}$ represents a linear projection from dimension $d$ to $c$; $\Phi_{\text{Proj-2}}$ is a linear projection within dimension $c$; $\Phi_{\text{GELU}}$ denotes the GELU activation function; and $c$ is the hidden layer dimension of the large language model.
% @@@@@@@@@@@@@@@@@@@@@@@@@@@@@@@@ Projector End @@@@@@@@@@@@@@@@@@@@@@@@@@@@@@@@

% >>>>>>>>>>>>>>>>>>>>>>>>>>>>>>>>> LLM <<<<<<<<<<<<<<<<<<<<<<<<<<<<<<<<<<

\subsection{Large Language Model}

We use a well-pretrained large language model to extract, decode, and generate descriptions of the detected changes. Specifically, we use the qwen2 series model~\cite{yang2024qwen2}.

The prompt used is as follows: "\texttt{This is Image1 <image1>. This is Image2 <image2>. What difference happened from Image1 to Image2?}” where \texttt{<image1>} and \texttt{<image2>} serve as placeholders for the images. The enhanced bi-temporal features, $\hat{F}_1 \in \mathbb{R}^{\frac{H}{14} \times \frac{W}{14} \times d}$ and $\hat{F}_2 \in \mathbb{R}^{\frac{H}{14} \times \frac{W}{14} \times d}$, are flattened along the token dimension and filled in at the corresponding positions.
The entire process can be represented as follows:
\begin{equation}
\begin{aligned}
E &= \Phi_{\text{tokenizer-emb}}(P) \\
T &= \Phi_{\text{llm-decode}}(\hat{F}_1, \hat{F}_2, E)) \\
\end{aligned}
\end{equation}

Here, $P$ represents the text portion of the prompt, $\Phi_{\text{tokenizer-emb}}$ is the tokenization and embedding stage, and $E$ is the encoded text portion. $\Phi_{\text{llm-decode}}$ denotes the decoding and text generation phase by the large language model, producing the change caption $T \in \mathcal{C}^{N}$, where $\mathcal{C}$ is the vocabulary of the generated caption.

% @@@@@@@@@@@@@@@@@@@@@@@@@@@@@@@@ LLM End @@@@@@@@@@@@@@@@@@@@@@@@@@@@@@@@

% >>>>>>>>>>>>>>>>>>>>>>>>>>>>>>>>> Train <<<<<<<<<<<<<<<<<<<<<<<<<<<<<<<<<<

\subsection{Training Strategy}

\subsubsection{Task Setup and Loss Function}

Given a training set $\mathcal{D}_{\text{train}} = {(\mathcal{I}_1, y_1), \dots, (\mathcal{I}_N, y_N)}$, where $\mathcal{I}_i = {\mathcal{I}_i^1 \in \mathbb{R}^{H \times W \times 3}, \mathcal{I}_i^2 \in \mathbb{R}^{H \times W \times 3}}$ represents a pair of bi-temporal remote sensing images, and $y_i \in \mathcal{C}^{N}$ denotes the corresponding change description annotation.

Our objective is to train a specially designed multimodal large model that can accurately generate change captions for any given image pair from a test set ($x_k \sim \mathcal{D}_{\text{test}}$).

The loss function we use is:
\begin{align}
    \begin{split}
        \mathcal{L} &= -\frac{1}{M}\sum_{j}^{M} w_j \log (\hat{w}_j) \\
    \end{split}
\end{align}
where $w_j$ and $\hat{w}_j$ represent the one-hot encoding and predicted word for the $j$-th word in the change description, respectively, and $M$ denotes the total number of words in the description.

\subsubsection{Stage-Wise Training Strategy} \label{sec:stage_training_strategy}

The pretraining of multimodal large models typically occurs in multiple stages, such as image-text alignment, large language model fine-tuning, and overall model fine-tuning. Given the introduction of a novel module, directly applying supervised fine-tuning (SFT) may disrupt the well-pretrained features. Therefore, our continued pretraining is divided into three stages, as illustrated in Fig.~\ref{fig:CCExpert_overall_arch}. Additionally, to support this hierarchical pretraining approach, we constructed the CC-Foundation Dataset, detailed in Section \ref{sec:dataset}.

In the first stage, only the "Difference Capture and Injection" and "Multi-Scale Adaptive Weighting and Adjustment" modules are trained, while the parameters of the image encoder and large language model are frozen. Using the CC-Foundation Dataset, this stage aims to provide a robust initialization for these modules. Following this pretraining phase, these modules can accurately extract differential features and inject them into the existing image features, which have strong generalization capabilities and are interpretable by the language model.

In the second stage, all model parameters are unfrozen, with the learning rate for the image encoder set to 0.2 times the overall learning rate. This stage is intended to improve the language model’s understanding of the image features and refine text generation, while simultaneously optimizing other modules to enhance performance in response to the improved language model.

In the third stage, all parameters remain unfrozen, with the image encoder’s learning rate still set to 0.2 times the overall rate. At this point, the model is trained on domain-specific data to ensure CCExpert achieves optimal performance in practical applications.
\section{Experimental Results and Analyses}

To validate the effectiveness of CCExpert, we conducted extensive experiments, which are organized into the following sections: dataset and experimental details, comparison with existing methods, ablation studies, and discussion.

\subsection{Experimental Datasets and Settings}

\subsubsection{CC-Foundation Dataset} \label{sec:dataset}

\begin{figure*}[!thb]
\begin{center}
\includegraphics[width=\linewidth]{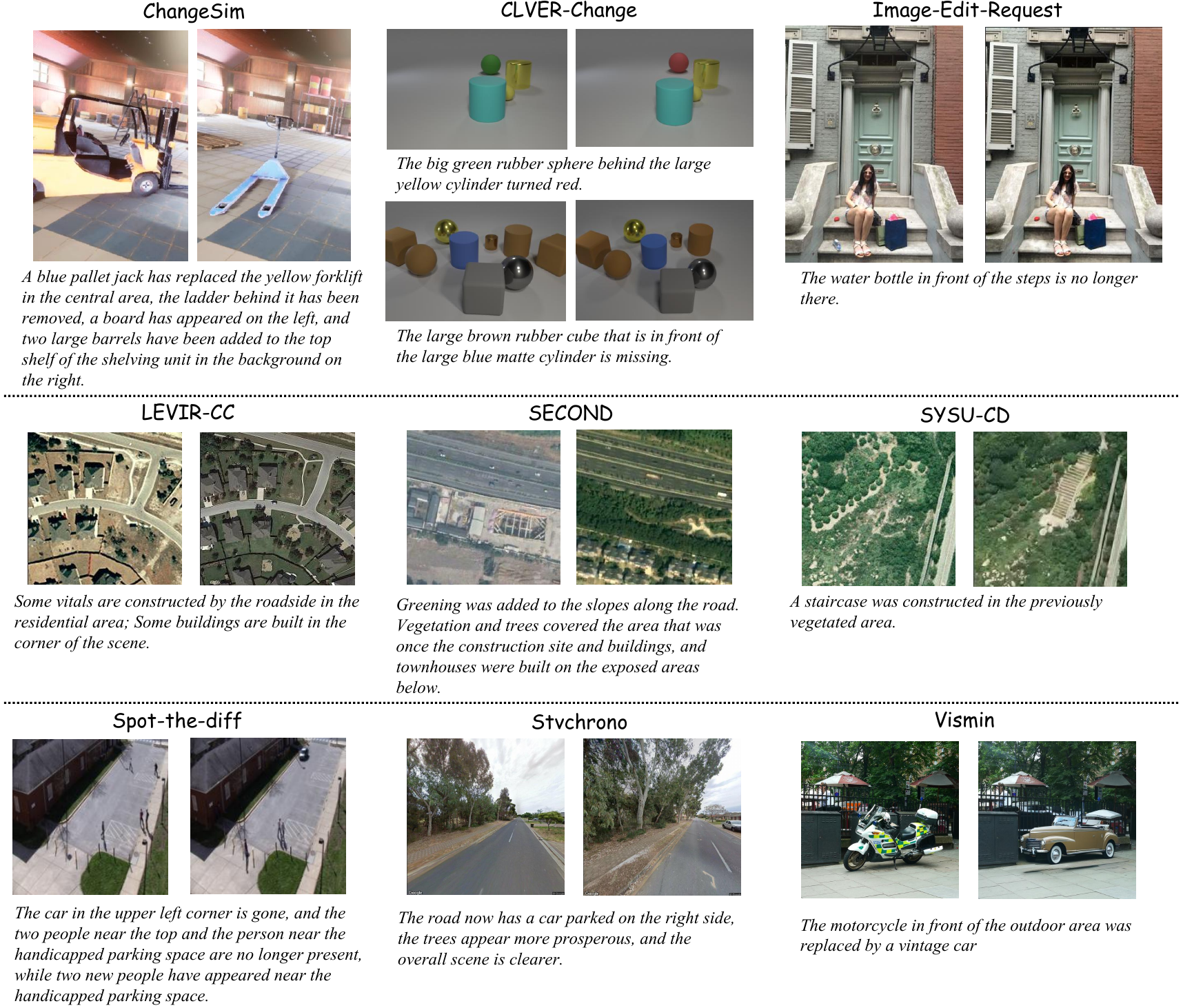}
\end{center}
\caption{
Visualization of the CC-Foundation Dataset.
}
\label{fig:cc_foundation_dataset}
\end{figure*}

In this paper, we introduce the CC-Foundation Dataset, a large-scale, content-rich dataset specifically curated for the Change Captioning domain. The data sources and quantities are detailed in Table~\ref{tab:dataset_tab}. Some subsets only use a portion of the available data; “Rewrite \& Refine” indicates whether the original annotations have been modified and refined, while “Annotation (Anno.) Type” indicates the source of the annotations. Sample visualizations are shown in Fig.~\ref{fig:cc_foundation_dataset}.

First, we included as many open-source datasets in this field as possible, such as CLVER-Change~\cite{park2019robust}, ImageEdit~\cite{tan2019expressing}, Spot-the-diff~\cite{jhamtani2018learning}, stvchrono~\cite{sun2024stvchrono}, Vismin~\cite{awal2024vismin}, and LEVIR-CC~\cite{RSICCformer}. We then utilized a large language model (LLM) to further refine these data, enhancing the annotations through tasks such as merging, rephrasing, language refinement, and enriching expression diversity.

Additionally, we incorporated datasets containing change detection masks, such as ChangeSim~\cite{park2021changesim} and SYSU-CD~\cite{shi21deeply}. The data were first resized based on the significance of the changes, after which we leveraged the generalization capabilities of GPT-4o~\cite{achiam2023gpt}, using the explicit information provided by the change masks to generate detailed change descriptions through multi-turn dialogue.

Finally, we introduced the SECOND~\cite{SECOND} dataset, which includes a variety of semantic change image pairs, such as changes in buildings, water bodies, and roads, along with finely annotated bi-temporal masks. Using GPT-4o, we generated a set of change prompts, followed by annotation work from domain experts. This addition expands the proportion of remote sensing imagery in the CC-Foundation Dataset, enhancing its diversity and level of challenge.

\begin{table}[!thb]
\centering
\caption{Overview of the CC-Foundation Dataset.
}
\label{tab:dataset_tab}
\resizebox{\linewidth}{!}{
\begin{tabular}{l | *{4}{c}}
\toprule
Dataset & Images & Captions & Rewrite\&Refine & Anno. type \\
\midrule 
CLVER-Change~\cite{park2019robust}                 & 40k & 493k  & \ding{55}  & Manual \\
Image-Edit-Request~\cite{tan2019expressing}        & 4k  & 12k   & \checkmark & Manual \\
Spot-the-diff~\cite{jhamtani2018learning}           & 12k & 62k   & \checkmark & Manual \\
Stvchrono~\cite{sun2024stvchrono}                  & 10k & 50k   & \checkmark & Manual \\
Vismin~\cite{awal2024vismin}                       & 83k & 420k  & \checkmark & Manual \\
LEVIR-CC~\cite{RSICCformer}                            & 10k & 50k   & \ding{55}  & Manual \\
ChangeSim~\cite{park2021changesim}                 & 24k & 72k   & -          & GPT4o \\
SYSU-CD~\cite{shi21deeply}                         & 13k & 65k   & -          & GPT4o \\
SECOND~\cite{SECOND}                               & 5k  & 20k   & \ding{55}  & Manual \\
\midrule 
ALL                                                & 20w & 120w  & -          & - \\
\bottomrule
\end{tabular}
}
\end{table}

\subsubsection{Evaluation Dataset}
To evaluate the effectiveness of the proposed method, we selected the LEVIR-CC~\cite{RSICCformer} dataset as the test set. LEVIR-CC is a large-scale remote sensing image captioning dataset containing 10,077 image pairs and 50,385 descriptive captions. These images depict changes across 20 different regions in Texas, USA, over a span of 5 to 15 years.

\subsubsection{Evaluation Protocol and Metrics}
We use four quantitative metrics to assess the accuracy of caption generation~\cite{RSICCformer}: BLEU-N (N=1,2,3,4), ROUGE$_L$, METEOR, and CIDEr-D.
Higher scores across these metrics indicate better model performance. Additionally, following~\cite{RSICCformer}, we employ the average metric $S^*_m$, defined as:
\begin{equation} 
S^*_m = \frac{1}{4}\mathrm{(BLEU\text{-}4+ROUGE_L+METEOR+CIDEr\text{-}D)} \\
\end{equation}

\subsubsection{Implementation Details}
Thanks to LLava-OneVision's~\cite{li2024llava} extensive training on image-text paired data, it already possesses strong capabilities in the Change Captioning domain. Consequently, the weights of the visual encoder, projector, and LLM in CCExpert are initialized from LLaVA-OneVision. Specifically, the visual encoder uses the siglip-400m~\cite{zhai2023sigmoid} model, the Projector consists of a two-layer MLP with interleaved GELU activation functions, and the LLM is based on the Qwen2 architecture~\cite{yang2024qwen2}. The custom-designed difference enhancement module uses multi-scale features from layers $\{-11, -8, -5, -2\}$ of the visual encoder, with each difference capture module comprising two stacked Change Aware Transformer Layers.

During training, following the input size used in Siglip-400m~\cite{zhai2023sigmoid} pretraining, CCExpert employs a fixed input image size of $384 \times 384$ for both training and testing, without any data augmentation. Supervised training uses binary cross-entropy loss with AdamW as the optimizer.

Unless otherwise noted, all experiments use a three-stage progressive training approach with the CC-Foundation dataset. LEVIR-CC, SYSU-CD, and SECOND datasets are upsampled threefold to increase the proportion of remote sensing images in training. The basic parameter configuration for the three-stage training process is provided in Table~\ref{tab:ccexpert_hyper_tab}, where "Data Iter" indicates the handling method for multiple change captions per image pair: "Flatten" refers to adding all unique captions to the dataset simultaneously, and "Random Choice" refers to randomly sampling one caption per iteration. Experiments were conducted on eight NVIDIA H100 GPUs, with the algorithm implemented on the Huggingface platform using PyTorch and DeepSpeed as the backend frameworks.

\begin{table}[!thb]
\centering
\caption{Hyperparameter Configuration for the Three-Stage Training of CCExpert
}
\label{tab:ccexpert_hyper_tab}
\resizebox{\linewidth}{!}{
\begin{tabular}{l | *{5}{c}}
\toprule
Stage & Batchsize & Epoch & Base LR & ViT-LR & Data Iter \\
\midrule 
Stage-I                 & 64 & 1  & 1e-5  & -  & Flatten \\
Stage-II                & 256 & 1  & 1e-5  & 2e-6  & Flatten \\
Stage-III               & 256 & 50  & 1e-5  & 2e-6  & Random Choice \\
\bottomrule
\end{tabular}
}
\end{table}

\subsection{Comparison with the State-of-the-Art}
The effectiveness of the CCExpert method was evaluated through comparisons with various state-of-the-art remote sensing image change captioning methods. The selected methods span a significant timeline and encompass a wide range of technical approaches. The following sections present the evaluation in terms of quantitative and qualitative metrics.

\subsubsection{Quantitative Comparisons}
The specific metrics for quantitative comparison on the LEVIR-CC dataset are shown in Table~\ref{tab:Comparisons_other_methods}. The performance metrics presented are based on either officially published results or reimplementations in PyTorch following the exact specifications provided in each respective paper. The proposed CCExpert model improves the overall performance metric $S^*_m$ by nearly 2 points compared to the current state-of-the-art, demonstrating that selecting the appropriate base model, coupled with the difference enhancement module and the staged pretraining strategy on the CC-Foundation dataset, significantly enhances multimodal model performance on this task. Notably, CCExpert excels in METEOR and CIDEr-D scores, outperforming all current comparison methods, which strongly validates CCExpert’s superior capability in generating diverse, accurate, and comprehensive text. This highlights its potential in the remote sensing image change captioning field.

\begin{table*}[!htbp]
\centering
\caption{Comparison of performance using various methods on LEVIR-CC test sets.}
\label{tab:Comparisons_other_methods}
\resizebox{\linewidth}{!}{
\begin{tabular}{c | c c c c c c c | c}
\toprule
Method & BLEU-1 & BLEU-2 & BLEU-3 & BLEU-4 & METEOR & ROUGE$_L$ & CIDEr-D & $S^*_m$ \\
\midrule
{Capt-Rep-Diff~\cite{robust_CC}} & 72.90 & 61.98 & 53.62 & 47.41 & 34.47 & 65.64 & 110.57 & 64.52\\
{Capt-Att~\cite{robust_CC}} & 77.64 & 67.40 & 59.24 & 53.15 & 36.58 & 69.73 & 121.22 & 70.17\\
{Capt-Dual-Att~\cite{robust_CC}} & 79.51 & 70.57 & 63.23 & 57.46 & 36.56 & 70.69 & 124.42 & 72.28\\
{DUDA ~\cite{robust_CC}} & 81.44 & 72.22 & 64.24 & 57.79 & 37.15 & 71.04 & 124.32 & 72.58\\
{MCCFormer-S~\cite{MCCformer}} & 79.90 & 70.26 & 62.68 & 56.68 & 36.17 & 69.46 & 120.39 & 70.68\\
{MCCFormer-D~\cite{MCCformer}} & 80.42 & 70.87 & 62.86 & 56.38 & 37.29 & 70.32 & 124.44 & 72.11\\
{RSICCFormer~\cite{RSICCformer}} & {84.72} & {76.27} & {68.87} & {62.77} & {39.61} & {74.12} & {134.12} & 77.65\\
{PSNet~\cite{PSNet}} & 83.86 & 75.13 & 67.89 & 62.11 & 38.80 & 73.60 & 132.62 & 76.78\\
{PromptCC~\cite{PromptCC}} & 83.66 & 75.73 & 69.10 & 63.54 & 38.82 & 73.72 & 136.44 & 78.13\\
{Sen~\cite{Sen}} & 85.10 & 77.05 & 70.01 & 64.09 & 39.59 & 74.57 & 136.02 & 78.57 \\
{SFT~\cite{Pix4Cap}} & 84.56 & 75.87 & 68.64 & 62.87 & 39.93 & 74.69 & 137.05 & 78.63 \\
{Pix4Cap~\cite{Pix4Cap}} & {85.56} & {77.08} & {69.79} & {63.78} & {39.96} & {75.12} & {136.76} & {78.91}\\
{Chg2Cap~\cite{Chg2Cap}} & 86.14 & 78.08 & 70.66 & 64.39 & 40.03 & 75.12 & 136.61 & 79.03\\
{RSCaMa~\cite{RSCaMa}} & 85.79 & 77.99 & 71.04 & 65.24 & 39.91 & 75.24 & 136.56 & 79.24\\
{ChangeChat~\cite{ChangeChat}} & 83.14 & - & -  & -  & 38.73 & 74.01 & 136.56 & - \\
{KCFI~\cite{KCFI}} & 86.34 & 77.31 & 70.89 & 65.30 & 39.42 & 75.47 & 138.25 & 79.61 \\
% {Semantic-CC~\cite{Semantic-CC}} & 88.07 & 79.68 & 71.47 & 64.51 & 40.58 & 77.76 & 138.51 & 80.34 \\

\midrule
{\textbf{CCExpert-0.5B}} & 86.41 & 78.31 & 71.30 & 65.42 & 41.33 & 75.93 & 141.19 & 80.99 \\
{\textbf{CCExpert-7B}} & \textbf{86.65} & \textbf{78.47} & \textbf{71.31} & \textbf{65.49} & \textbf{41.82} & \textbf{76.55} & \textbf{143.32} & \textbf{81.80} \\

\bottomrule%[1pt]
\end{tabular}
}
\end{table*}

\subsubsection{Qualitative Comparisons}

\begin{figure*}[!thb]
\begin{center}
\includegraphics[width=\linewidth]{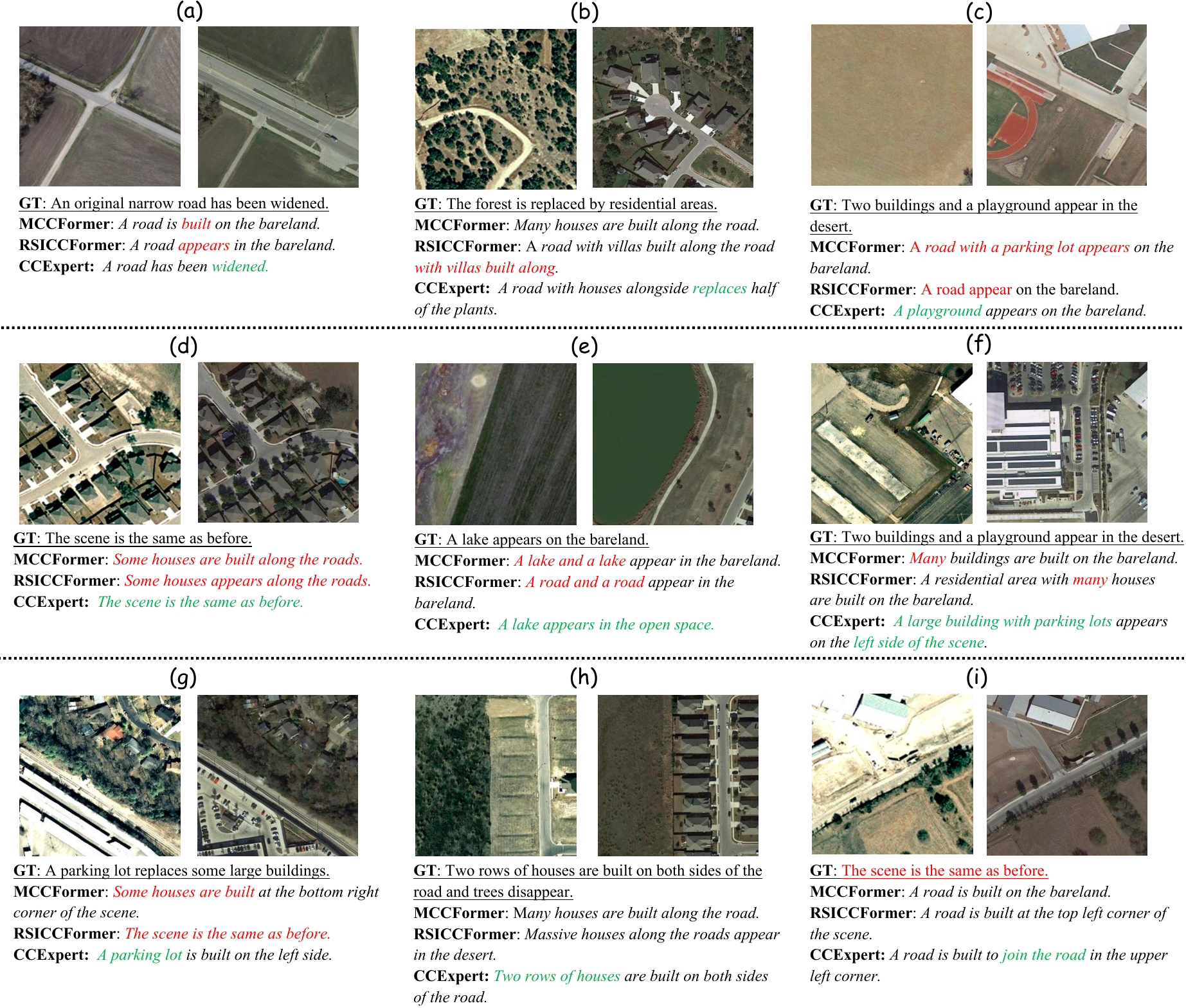}
\end{center}
\caption{
Visual comparisons of the proposed method with other state-of-the-art methods.
}
\label{fig:CCExpert_pred_results_compare}
\end{figure*}

We also visualized the prediction results of MCCFormer-D, RSICCFormer, and CCExpert on the LEVIR-CC dataset, as shown in Fig.~\ref{fig:CCExpert_pred_results_compare}. Incorrect predictions are marked in red, while correct predictions are marked in green.

\begin{itemize}
    \item (a) illustrates CCExpert’s advantage in feature comparison. CCExpert accurately captures the key information of a narrow road being widened, closely matching the GT term “widened.” In contrast, other methods generate the more common dataset phrase, “A road is built on the bareland,” failing to recognize the subtle differences in road features across the two images.

    \item (b) shows CCExpert’s ability to retain the context of the original image, whereas other methods only focus on the post-change form, lacking global context.

    \item (c) demonstrates CCExpert’s precision in extracting visual information, accurately identifying the key entity “playground,” while other methods make errors, such as misidentifying it as a “road” or “parking lot.” Similarly, (g) confirms CCExpert’s advantage in key entity recognition.

    \item (d) reflects CCExpert’s robust ability to filter irrelevant noise and extract essential features, while (g) and (i) show how overexposure in images can impact predictive accuracy.

    \item (e) highlights CCExpert’s advantage in language generation, consistently producing grammatically accurate descriptions due to the conversational capabilities of the powerful LLM (large language model).

    \item (f) demonstrates CCExpert’s excellence in differential analysis and refined expression, as well as its spatial positioning capabilities.

    \item (h) and (i) indicate CCExpert’s high sensitivity to detail and its ability to capture human language preferences. For instance, terms like “two columns” enhance the natural flow of the description, aligning with human linguistic tendencies.
\end{itemize}

In summary, CCExpert’s performance showcases its superior capabilities in visual feature extraction, semantic change recognition with noise filtering, spatial positioning, and natural language generation.

\subsection{Ablation Study}
This section aims to verify the effectiveness of the proposed components and strategies through a series of ablation experiments. Specifically, it examines the impact of the base model selection, the effectiveness of the CC-Foundation Dataset, the performance of the difference enhancement module, and specific parameter choices.

\subsubsection{Effects of Base Model}

\begin{table*}[!thb]
\centering
\caption{The effects of different base model. Performances are validated on LEVIR-CC test set.}
\label{tab:CCExpert_Basemodel}
\resizebox{0.7\linewidth}{!}{
\begin{tabular}{c c | c c c c | c}
\toprule
Base Model & Model Size & BLEU-4 & METEOR & ROUGE$_L$ & CIDEr-D & $S^*_m$ \\
\midrule
MiniGPT4-7B~\cite{minigpt4} & 1.1B+7B & 62.74 & 39.69 & 74.46 & 136.13 & 78.25 \\
LLaVA-1.5-7B~\cite{liu2023llava} & 0.3B+7B & 62.89 & 39.90 & 75.29 & 136.76 & 78.71 \\
LLaVA-ov-0.5B~\cite{li2024llava} & 0.4B+0.5B & \textbf{64.53} & 40.73 & 75.03 & 138.01 & 79.58 \\
LLaVA-ov-7B~\cite{li2024llava} & 0.4B+7B & 63.59 & \textbf{41.45} & \textbf{75.93} & \textbf{139.80} & \textbf{80.19} \\
\bottomrule%[1pt]
\end{tabular}
}
\end{table*}

In recent years, multimodal large models (MLLMs) have demonstrated exceptional performance across various downstream tasks due to their strong generalization capabilities. The choice of base multimodal model is a critical factor that directly impacts performance on these tasks. Therefore, we initially fine-tuned several different MLLMs on the LEVIR-CC dataset under the same configuration; detailed metrics are presented in Table~\ref{tab:CCExpert_Basemodel}.

In this study, LLaVA-ov emerged as the most effective model, attributable to its extensive image-text pretraining data and the robust integration of its vision branch and language model components. Even with a base language model size of only 0.5B parameters, LLaVA-ov-0.5B significantly outperformed LLaVA-1.5-7B. Moreover, with the increase in language model parameters, LLaVA-ov-7B excelled on the LEVIR-CC benchmark, even without additional data or specialized components. Based on these results, we selected LLaVA-ov as the foundational model for further research.

\subsubsection{Effects of Continue Pretraining Dataset}

\begin{table*}[!thb]
\centering
\caption{The effects of continue pretraining dataset: CC-Foundation Dataset. Performances are validated on LEVIR-CC test set.}
\label{tab:CCExpert_Data}
\resizebox{\linewidth}{!}{
\begin{tabular}{c c | c c c c c c c | c c}
\toprule
Method & CC-Foundation & BLEU-1 & BLEU-2 & BLEU-3 & BLEU-4 & METEOR & ROUGE$_L$ & CIDEr-D & $S^*_m$ & $\Delta S^*_m$\\
\midrule
\multirow{2}{*}{\shortstack{CCExpert-0.5B}} & \ding{55}   & 85.84 & 77.51 & 70.34 & 64.52 & 40.73 & 75.03 & 138.01 & 79.58 & -   \\
                                   & \checkmark  & \textbf{86.49} & \textbf{77.97} & \textbf{70.57} & \textbf{64.53} & \textbf{41.20} & \textbf{75.64} & \textbf{140.99} & \textbf{80.49} & \textbf{+0.91}\\
\midrule
\multirow{2}{*}{\shortstack{CCExpert-7B}}   & \ding{55}   & 86.37 & 77.67 & 69.94 & 63.59 & 41.45 & 75.93 & 139.80 & 80.19 & -  \\
                                   & \checkmark  & \textbf{86.35} & \textbf{77.90} & \textbf{70.50} & \textbf{64.32} & \textbf{41.60} & \textbf{76.34} & \textbf{142.98} & \textbf{81.31} & \textbf{+1.12}  \\

\bottomrule%[1pt]
\end{tabular}
}
\end{table*}
Although LLaVA-ov easily outperforms other models on the benchmark, we believe there is room for further optimization. Following the application paradigm for large models and leveraging the CC-Foundation Dataset, we conducted continued pretraining for both the 0.5B and 7B versions. This continued pretraining process follows the three-stage approach presented in Section~\ref{sec:stage_training_strategy}, with specific metrics shown in Table~\ref{tab:CCExpert_Data}.

After continued pretraining with the CC-Foundation Dataset, the 0.5B version of CCExpert improved its $S^*_m$ score from 79.58 to 80.49, an increase of 0.91. The 7B version showed an even greater improvement, from 80.19 to 81.31, with a gain of 1.12. All evaluation metrics improved, and the model’s performance after continued pretraining was significantly enhanced. These results indicate that continued pretraining on high-quality, domain-specific data is an effective way to improve model performance on specific tasks.

\subsubsection{Effects of Difference-aware Integration module}

\begin{table*}[!thb]
\centering
\caption{The effects of dfifference explicit injection module(Enhancer). Performances are validated on LEVIR-CC test set. }
\label{tab:CCExpert_model_design}
\resizebox{\linewidth}{!}{
\begin{tabular}{c c | c c c c c c c | c c}
\toprule
Model Size & Enhancer & BLEU-1 & BLEU-2 & BLEU-3 & BLEU-4 & METEOR & ROUGE$_L$ & CIDEr-D & $S^*_m$ & $\Delta S^*_m$\\
\midrule
\multirow{2}{*}{\shortstack{CCExpert-0.5B}} & \ding{55}   & \textbf{86.49} & 77.97 & 70.57 & 64.52 & 41.20 & 75.64 & 140.99 & 80.49 & -   \\
                                   & \checkmark  & 86.41 & \textbf{78.31} & \textbf{71.30} & \textbf{65.42} & 41.33 & \textbf{75.93} & \textbf{141.19} & \textbf{80.99} & +0.50\\
\midrule
\multirow{2}{*}{\shortstack{CCExpert-7B}}   & \ding{55}   & 86.35 & 77.90 & 70.50 & 64.32 & 41.60 & 76.34 & 142.98 & 81.31 & -  \\
                                   & \checkmark  & \textbf{86.65} & \textbf{78.47} & \textbf{71.31} & \textbf{65.49} & \textbf{41.82} & \textbf{76.55} & \textbf{143.32} & \textbf{81.80} & +0.49  \\
\bottomrule%[1pt]
\end{tabular}
}
\end{table*}

Table~\ref{tab:CCExpert_model_design} presents the specific performance gains achieved by adding the Difference-aware Integration module after continued pretraining.

Since introducing this module directly during fine-tuning without pretraining could potentially impact the pretrained parameters of the base model, we did not perform experiments with this initialization approach. Instead, we compared the baseline performance after three-stage training with and without the Difference-aware Integration module.

For the CCExpert-0.5B version, introducing the Difference-aware Integration module increased $S^*_m$ from 80.49 to 80.99, a gain of 0.50. All metrics improved, with BLEU-4, ROUGE$_L$, and CIDEr-D rising to 65.42, 75.93, and 141.19, respectively.

For the CCExpert-7B version, the module further enhanced overall performance, with $S^*_m$ increasing from 81.31 to 81.80, a gain of 0.49. BLEU, METEOR, ROUGE$_L$, and CIDEr-D all reached optimal values, with CIDEr-D achieving 143.32, underscoring the module’s advantages in capturing visual details and enhancing change region features.

These results indicate that the Difference-aware Integration module significantly boosts performance for both the 0.5B and 7B versions of CCExpert, validating its effectiveness in enhancing model performance.

\subsubsection{Effects of Using Different Multi-Scale Features in the Difference-aware Integration Module}

\begin{table}[!thb]
\centering
\caption{The effects of using different multi-scale features in the Difference-aware Integration module. Performances are validated on LEVIR-CC test set.}
\label{tab:CCExpert_multi_level_features}
\resizebox{\linewidth}{!}{
\begin{tabular}{c | c c c c | c}
\toprule
Multi-scale features From & BLEU-4 & METEOR & ROUGE$_L$ & CIDEr-D & $S^*_m$ \\
\midrule
{$\{-2\}$} & 63.81 & 41.26 & 76.17 & 142.77 & 81.00 \\
{$\{-2,-4,-6,-8\}$}  & 64.13 & \textbf{42.20} & 76.39 & 143.40 & 81.53 \\
{$\{-2,-5,-8,-11\}$} & \textbf{65.49} & 41.82 & \textbf{76.55} & \textbf{143.32} & \textbf{81.80} \\
{$\{-2,-6,-10,-14\}$} & 64.45 & 41.03 & 75.89 & 140.34 & 80.43 \\
\bottomrule%[1pt]
\end{tabular}
}
\end{table}

In designing the difference enhancement module, our main objective was to inject change information into the image representation. However, certain change features may become less prominent as the ViT layers deepen. Thus, we experimented with extracting and integrating multi-scale features at different levels. For this purpose, we used the CCExpert-7B version and selected various sets of visual branch outputs as multi-scale features. The comparison results are presented in Table~\ref{tab:CCExpert_multi_level_features}

When using only a single-layer feature at $\{-2\}$, CCExpert showed relatively lower performance across all metrics, especially for CIDEr-D and $S^*_m$.

With features from layers $\{-2, -4, -6, -8\}$, the METEOR score significantly improved to 42.20, with slight gains in CIDEr-D and $S^*_m$ as well.

When using features from layers $\{-2, -5, -8, -11\}$ as the multi-scale feature input, CCExpert achieved optimal performance on BLEU-4, ROUGE$_L$, and $S^*_m$, with CIDEr-D also nearing the highest score at 143.32. This suggests that fusing features extracted at various depths substantially enhances overall model performance.

However, when the layer intervals were further increased to $\{-2, -6, -10, -14\}$, performance declined. Although CIDEr-D remained relatively high, $S^*_m$ and other metrics showed varying degrees of reduction.

Based on these results, we selected $\{-2, -5, -8, -11\}$ as the optimal layer combination for multi-scale feature input.

\subsubsection{Effects of Using Different Numbers of Change Aware Transformer Layers in the Diff Expert Module}

\begin{table}[!thb]
\centering
\caption{The effects of using different numbers of change aware transformer layers in the diff expert module. Performances are validated on LEVIR-CC test set.}
\label{tab:Comparisons_other_methods}
\resizebox{\linewidth}{!}{
\begin{tabular}{c | c c c c | c}
\toprule
Layers & BLEU-4 & METEOR & ROUGE$_L$ & CIDEr-D & $S^*_m$ \\
\midrule
1 & 63.90 & 41.78 & 76.06 & 141.79 & 80.90 \\
2 & \textbf{65.49} & 41.82 & \textbf{76.55} & \textbf{143.32} & \textbf{81.80} \\
3 & 65.32 & 41.67 & 76.42 & 143.04 & 81.61 \\
\bottomrule%[1pt]
\end{tabular}
}
\end{table}

Additionally, we conducted comparative experiments using different numbers of Change Aware Transformer Layers in the Diff Expert module with the CCExpert-7B version, as shown in Table~\ref{tab:CCExpert_multi_level_features}.

With only one Change Aware Transformer Layer, the model performed relatively poorly across all metrics, with $S^*_m$ reaching only 80.90, even falling below the baseline. This suggests that a single Change Aware Transformer Layer is insufficient for effectively capturing differential information.

When using two Change Aware Transformer Layers, the model achieved optimal performance across all metrics, with $S^*_m$ reaching 81.80. At two layers, the Diff Expert effectively captures changes and injects relevant information, significantly improving text generation quality.

However, when the number of layers increased to three, performance slightly declined. Although CIDEr-D remained high (143.04), BLEU-4 and $S^*_m$ showed small decreases. This indicates that in multi-layer configurations, too many Transformer layers may lead to feature redundancy or excessive information blending, ultimately impacting model performance.

\subsection{Discussions}

In this study, we propose an innovative multimodal large model-based approach, CCExpert, for generating change captions for remote sensing images. This method integrates the construction and application of continued pretraining data and emphasizes the importance of explicitly injecting differential features into the context, enabling the model to capture subtle visual changes between bi-temporal images accurately. The framework is designed to deeply understand change information and generate text descriptions that are highly consistent with visual changes.

Experimental results show that continued pretraining with the CC-Foundation dataset significantly improves the model’s performance across multiple metrics. Both the base version (0.5B) and the large version (7B) of CCExpert demonstrate consistent gains on metrics such as BLEU and CIDEr-D, underscoring the importance of domain-specific pretraining data in enhancing the model’s performance on specialized tasks. Furthermore, the Difference-aware Integration module excels in enriching change-sensitive features, providing the LLM with robust visual information that captures multi-scale changes, resulting in more accurate and detailed text descriptions.

Despite CCExpert’s strong performance, several limitations and potential future directions remain:

i) This study explores model architecture design and continued pretraining based on the LLaVA-ov-0.5B and 7B versions. Utilizing even larger-scale models in the future may further improve feature extraction and generalization capabilities. Future work could investigate larger base models, such as a 72B multimodal large model.

ii) The training data primarily comes from open datasets, which may contain unfiltered noise. Additionally, some data annotations were generated using GPT-4o based on change-mask prompts, intended to distill some of the capabilities of established multimodal large models but potentially containing errors. To enhance CCExpert’s generalization and predictive accuracy, future work could extend and curate high-quality, cleaned datasets, supplementing them with manually annotated data to further boost model performance.

iii) Due to the limited availability of benchmarks for remote sensing image change captioning, this study validates CCExpert only on the LEVIR-CC dataset, which may impose some limitations. We hope to expand the benchmarks in the future, potentially by extracting and curating a portion of the manually annotated SECOND subset from the CC-Foundation dataset as a benchmark. This remains a promising avenue for future research.

\section{Conclusion}

This paper presents an innovative approach for RSICC, called CCExpert. CCExpert builds upon an advanced multimodal large model architecture and introduces a Difference-aware Integration Module, which captures multi-scale differences in images and integrates them into the original image context, thereby enhancing the signal-to-noise ratio of salient features.

Additionally, we constructed the CC-Foundation Dataset, comprising 200,000 image pairs and 1.2 million captions, providing extensive data support for continued pretraining within this domain. Through a three-stage training process, the Difference-aware Integration Module is deeply integrated with the MLLM, significantly enhancing model performance in remote sensing image change captioning tasks.

Experimental results demonstrate that CCExpert achieves state-of-the-art performance on the LEVIR-CC benchmark, with a significant gap over other methods.

\ifCLASSOPTIONcaptionsoff
  \newpage
\fi

\small{
\bibliographystyle{IEEEtran}
\bibliography{IEEEabrv,myreferences}
}

\end{document}